\documentclass[letterpaper, 10 pt, conference]{ieeeconf}  
\usepackage[utf8]{inputenc}

\IEEEoverridecommandlockouts                              

\usepackage{cite}
\usepackage{amsmath,amssymb,amsfonts}
\usepackage{algorithmic}
\usepackage{graphicx}
\usepackage{textcomp}
\usepackage{xcolor}
\usepackage{bm}
\usepackage{balance}
\usepackage{multirow}
\usepackage{makecell}
\usepackage{booktabs}
\usepackage{threeparttable}
\usepackage{pifont}
\usepackage{soul}
\usepackage[pagewise]{lineno}
\usepackage{comment}

\soulregister{\cite}7
\soulregister{\ref}7
\soulregister{\ding}7
\usepackage[ruled,linesnumbered]{algorithm2e}

\title{\LARGE \bf
Collision-Aware Object-Goal Visual Navigation via Two-Stage Deep Reinforcement Learning}

\author{Hongwu Wang, Shiwei Lian and Feitian Zhang*
\thanks{*This work was not supported by any organization}
\thanks{All the authors are with the Robotics and Control Laboratory, School of Advanced Manufacturing and Robotics, and the State Key Laboratory of Turbulence and Complex Systems, Peking University, Beijing, 100871, China (emails: whw@stu.pku.edu.cn, {lianshiwei@stu.pku.edu.cn} and {feitian@pku.edu.cn}).}%
\thanks{* Send all correspondence to Feitian Zhang.}
}

\begin{document}
\maketitle

\thispagestyle{empty}
\pagestyle{empty}

\begin{abstract}
Object-goal visual navigation aims to reach a specific target object using egocentric visual observations. Recent deep reinforcement learning (DRL) approaches have achieved promising success rates but often neglect collisions during evaluation, limiting real-world deployment. To address this issue, this letter introduces a collision-aware evaluation metric, namely collision-free success rate (CF-SR), to explicitly measure navigation performance under collision constraints. In addition, collision-free success weighted by path length (CF-SPL) is adopted to further evaluate navigation efficiency. Furthermore, a two-stage DRL training framework with collision prediction is proposed to improve collision-free navigation performance. In the first stage, a collision prediction module is trained by supervising the agent's collision states during exploration. In the second stage, leveraging the trained collision prediction, the agent learns to navigate toward target objects while avoiding collision. Extensive experiments across multiple navigation models in the AI2-THOR environment demonstrate consistent improvements in both CF-SR and CF-SPL. Real-world experiments further validate the effectiveness and generalization capability of the proposed framework.
\end{abstract}


\section{Introduction}
Object-goal visual navigation enables robots to reach  specified target objects using only egocentric visual observations. This capability is essential for various robotic applications, including service robotics, warehouse automation, and household assistance. Recent deep reinforcement learning (DRL) methods have demonstrated promising success rates using RGB inputs \cite{tdanet, apexnav25, emkg26, osmag26, SAVN, VTNet, OMT, ORG, hoz, akgvp, mjol, ral}. Some models implicitly learn visual representations \cite{SAVN, VTNet, OMT}, while others leverage object relationships, semantic memory, or open-vocabulary scene priors through knowledge graphs and language-grounded semantic maps \cite{ORG, hoz, akgvp, apexnav25, emkg26, osmag26} or context-aware reasoning \cite{mjol, ral, tdanet, glasses26} to improve navigation performance.

Despite these advances, many existing approaches still prioritize target-reaching success while paying limited attention to the safety of the navigation process, leaving collision avoidance insufficiently explored and hindering reliable deployment in real-world robotic scenarios. Existing evaluation metrics typically consider an episode successful if agent reaches the target, regardless of collisions encountered during navigation. As a result, learned policies may frequently collide with obstacles while still achieving high success rates. Directly penalizing collisions in reward design often leads to overly conservative policies and reduces exploration efficiency, thereby limiting overall navigation performance \cite{SAVN, VTNet, OMT, ORG, hoz, akgvp, mjol, ral, tdanet}. This limitation becomes particularly critical in real-world applications, where collisions could damage robots, surrounding objects, or the environment.

To address the aforementioned challenge, this letter introduces a collision-aware evaluation metric, collision-free success rate (CF-SR), to assess whether the agent reaches the target without collisions. Additionally, collision-free success weighted by path length (CF-SPL) is adopted to evaluate navigation efficiency. Building on these metrics, this letter further proposes a two-stage training framework with a collision prediction module. In the first stage, the agent explores freely while the collision predictor learns to estimate collision risk through self-supervised signals. In the second stage, the trained predictor guides navigation under a collision penalty, enabling the agent to learn collision-free paths while maintaining effective exploration. 

The contributions of this work are threefold.
First, this letter introduces a collision-aware evaluation metric, CF-SR, to quantify navigation performance under collision constraints, and adopts CF-SPL to evaluate navigation efficiency. Second, a two-stage DRL training framework with a collision prediction module is proposed to improve safety without compromising success rate. Third, extensive experiments on multiple navigation models in both simulation and real-world environments demonstrate consistent improvements in CF-SR and CF-SPL.

\section{Related Work}

\begin{figure*}[t]
\centerline{\includegraphics[width=0.82\linewidth]{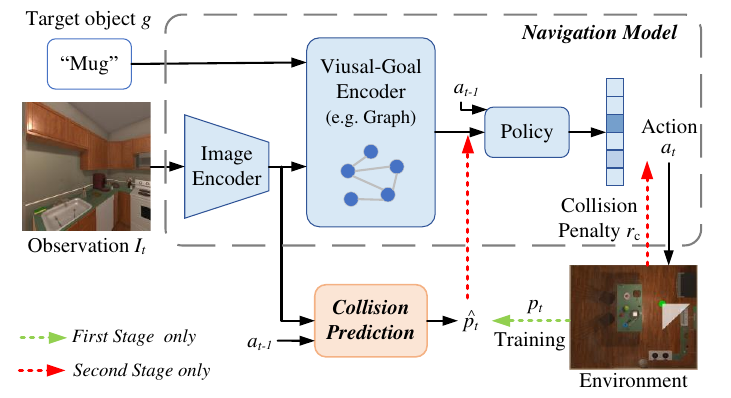}}
\caption{Collision-aware navigation framework with two-stage training and collision prediction.}
\label{fw}
\end{figure*}

\subsection{Object-Goal Visual Navigation}

Prior approaches for object-goal visual navigation are primarily classified into map-based and end-to-end methods. Map-based methods incrementally construct semantic maps \cite{cow, goat, roa, DBLP} or topological maps \cite{wu2024voronav, Semi} of the environment during exploration, requiring precise localization and depth sensing. End-to-end models directly map image and goal embeddings to actions using DRL. Many end-to-end studies utilize only RGB observations and achieve remarkable navigation performance with various network designs, including the meta learning\cite{SAVN}, visual transformer\cite{VTNet, OMT}, knowledge graph\cite{ORG, hoz, akgvp, DBLP3, 3d}, context information\cite{mjol, ral, SSNet, Li, imgo, OBJE}, layout information\cite{lstde}, attention mechanisms\cite{SpatialAtt, tdanet, vnsa}, and subtask learning\cite{tro}. For example, Du et al.\cite{VTNet} employed a visual transformer\cite{transformer} to extract informative visual representations during navigation, and Xu et al.\cite{akgvp} aligned the knowledge graph with visual perception via CLIP \cite{CLIP} to enhance navigation performance.

While these methods improve target-reaching success, they do not explicitly account for collisions along the trajectory. Consequently, their navigation performance may degrade when the collision is taken into account when calculating the success rate. To address this limitation, we propose a new metric, collision-free success (CF-SR), to evaluate the ability of navigation agents to reach the target without collisions.

\subsection{Collision Avoidance of DRL Visual Navigation}
Collision avoidance is essential for robot navigation. A common strategy is reshaping the reward function with a collision penalty\cite{humanoid, drqn,coll1, Yue, drl}. For instance, Xiao et al.\cite{coll1} introduced a single-step collision penalty to discourage collisions, however, this often produces overly conservative navigation policies, hindering the improvement of the success rate. Wu et al.\cite{dual} predicted the collision in advance as an auxiliary task of imitation learning, improving safety during navigation, but requiring expert trajectories for training. 

Our method differs in two key aspects. First, it does not require expert demonstrations. Second, it integrates a collision predictor into a two-stage DRL framework. In the first stage, the agent explores freely while the predictor learns to estimate collision risk. In the second stage, the predictor informs navigation under a collision penalty, enabling collision-free paths without compromising exploration efficiency. Experimental results demonstrate that our proposed method outperforms  conventional reward-based and auxiliary prediction methods in improving collision-free navigation.

\section{Task Definition}
In this letter, the agent is tasked to navigate to a set of predefined target objects $G$ using only egocentric RGB observations. Each episode begins with a randomly initialized agent position. At each time step $t$, the agent selects an action $a_t \in \mathcal{A}$ based on the current RGB image $I_t$ and the target object $g\in G$ by employing a policy network $\pi\left(a_t \mid I_t, g; \theta \right)$, where $\theta$ denotes the network parameters. The action space $\mathcal{A}$ is discretized as \{\texttt{MoveAhead}, \texttt{RotateLeft}, \texttt{RotateRight}, \texttt{LookUp}, \texttt{LookDown}, \texttt{Done}\}. The step size of \texttt{MoveAhead} is 0.25\,m. The agent rotates by $45^\circ$ and tilts the camera by $30^\circ$ per action. The \texttt{Done} action terminates the navigation episode.

Conventionally, an episode is considered successful if the agent executes \texttt{Done} while the target object is \texttt{visible}, which means the target object appears in the agent's current RGB image and within a distance of 1.5\,m. We extend this definition by introducing the collision-free success metric, i.e., an episode is considered successful only if the above conditions are met and no collision occurs during navigation. Specifically, we propose the collision-free success rate (CF-SR) as the primary metric. CF-SR measures the fraction of episodes in which the agent reaches the target without collisions, defined as 
\begin{equation}
\text{CF-SR} = \frac{1}{N}\sum^{N}_{i=1}S_i
\end{equation}
\noindent where $N$ is the total number of episodes and $S_i$ is the collision-free success indicator of the $i$-th episode with 1 representing success and 0 otherwise. Additionally, CF-SPL is adopted to evaluate navigation efficiency by incorporating path optimality into collision-free success, calculated as ${\frac{1}{N}\sum^{N}_{i=1} S_i \frac{d_{i}^*}{\max(d_{i}^*, d_i)}}$ where $d_i$ and $d_{i}^*$ represent the agent's path length and the shortest path length from its initial position to the target object in the $i$-th episode, respectively. 
This formulation provides a more realistic assessment of navigation performance and motivates the design of the proposed training method to improve collision-free success.

\section{Collision-Aware Navigation Framework}

\subsection{Framework Overview and Policy Formulation}
The proposed two-stage training framework enhances existing navigation models by integrating a collision prediction module, as illustrated in Fig.~\ref{fw}. A typical navigation model contains an image encoder to process RGB observations and a visual-goal encoder to encode features of the current observation and the target object $g$. All the features processed by encoders are then concatenated and passed through a policy network $\pi$ to predict the action, often utilizing several long-short term (LSTM) layers and fully connected networks (FCN). At time $t$, the collision prediction module estimates the probability $\hat{p}_t$ of a collision when the agent moves forward from the current RGB observation $I_t$, i.e., $\hat{p}_t=P(I_t;\phi)$, where $\phi$ represents the module parameters. The agent then selects its action based on the current RGB observation $I_t$ and target object $g$ as well as the predicted collision probability $\hat{p}_t$, i.e., 
\begin{equation}
\pi(a_t\mid I_t,g, \hat{p}_t;\theta)
\end{equation}
where $\theta$ is the weight of the policy network $\pi$. 

\subsection{Collision Prediction Module}
The collision prediction is treated as a binary classification problem, since only the \texttt{MoveAhead} action may cause collisions. The designed collision prediction module contains two LSTM layers with a hidden state size of 512. At time $t$, it takes the current RGB observation $I_t$ and the previous action $a_{t-1}$ as the input and outputs the predicted collision probability $\hat{p}_t$. The LSTM architecture is adopted to capture temporal dependencies between consecutive observations, which improves collision prediction accuracy in partially observable environments. The collision prediction module is trained by supervising the navigation trajectories. At each time step where the agent takes the \texttt{MoveAhead} action, the module receives a binary collision state $p_t$ with 1 denoting collision from the environment and updates its weight by minimizing the cross-entropy loss $L_{\rm c}$ as  
\begin{equation}\label{eq_lc}
    L_{\rm c}=-p_t\log\hat{p}_t-(1-p_t)\log(1-\hat{p}_t)
\end{equation}
This design enables the module to estimate collision risk from visual cues, providing actionable guidance for the policy network.

\subsection{Two-Stage Training Strategy}

We adopt a two-stage training strategy to decouple target-reaching behavior from collision avoidance, enabling effective exploration while progressively incorporating safety awareness.
In the first training stage, the agent is encouraged to fully explore the environments to find the target without explicitly considering collisions. The reward function includes no collision penalty and is the same as the original reward $r_{\rm o}$ that varies from the selected navigation models. The collision prediction module concurrently learns to predict collision probability when the agent takes the \texttt{MoveAhead} action by supervising collisions along the agent's trajectories. The input of collision probability into the policy network is set to zero in this training stage. The first stage encourages a bold exploration policy that maximizes target-reaching success without penalizing collisions.

\begin{algorithm}[t]
\caption{Per-Thread Two-Stage Training With Collision Prediction.}\label{algorithm}

\For{{\rm episode} $E = 1$ {\rm to} $E_1+E_2$}{
Randomize the scene, agent position and goal $g$\;
Reset time step $t\leftarrow0$\;
    \Repeat{$a_t=$ \texttt{Done}}{
    Reset gradients $d\theta\leftarrow 0$, $d\theta_v\leftarrow 0$, $d\phi\leftarrow 0$\;
    Synchronize $\theta'\leftarrow\theta$, $\theta_v'\leftarrow\theta_v$, $\phi'\leftarrow\phi$\;
    $t_{\rm start}\leftarrow t$\;
    \Repeat{$a_t=$ \texttt{Done} {\rm or} $t-t_{\rm start}=t_{\rm max}$}{
        Acquire image $I_t$\;
        Predict collision $\hat{p}_t\leftarrow P(I_t;\phi')$\;
        \eIf{$E< E_1$}
        {$O_t\leftarrow (I_t,g,0)$\;
        Execute $a_t$ through $\pi(a_t\mid O_t;\theta')$\;
        Observe collision state $p_t$ and  reward $r_t\leftarrow r_{{\rm o},t}$\;}
        {$O_t\leftarrow(I_t,g,\hat{p}_t)$\;
        Execute $a_t$ through $\pi(a_t\mid O_t;\theta')$\;
        Observe  collision state $p_t$ and  reward $r_t\leftarrow r_{{\rm o},t}+r_{{\rm c},t}$\;}
        $t\leftarrow t+1$\;
        }
    $R\leftarrow 0$ if $a_t=$ \textit{\texttt{Done}} else $V(O_t;\theta_v')$\;
    \For{$i\in\{t-1,...,t_{\rm start}\}$}{
        $R\leftarrow r_i+\gamma R$\;
        $\delta_t\leftarrow R-V(O_i;\theta_v')$\;
        $d\theta\leftarrow d\theta+\nabla_{\theta'}\log\pi(a_i\mid O_i;\theta')\delta_t$\;
        $d\theta_v \leftarrow d\theta_v +\nabla_{\theta_v'}\delta_{t}^{2}$\;
        \If{$E< E_1$ {\rm and} $a_i=$ \texttt{MoveAhead}}
        {
        $d\phi\leftarrow d\phi+\nabla_{\phi'} L_{{\rm c},i}$ using Eq.~(\ref{eq_lc})\;}
    }
    Perform asynchronous update of $\theta$, $\theta_v$ and $\phi$\;
    }
    
}
\end{algorithm}

In the second training stage, the agent learns to navigate to the target while avoiding collisions with the integration of the trained collision predictor. A collision penalty $r_{c}$ is introduced into the reward function. The predicted collision probability $\hat{p}_t$ learned from the first stage is inputted into the navigation policy to identify potential collisions. The parameters of the collision prediction module are fixed during this stage to prevent overfitting. This design enables the agent to balance exploration and safety, improving the CF-SR without excessively conservative behavior.

\subsection{Reward Function}
The reward at time step $t$ combines the original navigation reward $r_{\rm o}$ and the collision penalty $r_{\rm c}$, i.e.,
\begin{equation}
r = \left \{
\begin{array}{ccl}
r_{\rm o}  &      &  E<E_1\\
r_{\rm o}+r_{\rm c}   &      & E_1\leq E< E_1+E_2
\end{array} \right.
\end{equation}
\noindent Here,  $E$ is the global episode counter, $E_1$ and $E_2$ are the total numbers of episodes in the first and second training stages, respectively. $r_{\rm c}$ is set to -0.5 when the agent collides with obstacles, and 0 otherwise. Typical settings for $r_{\rm o}$, depending on specific navigation models, reward the agent for successfully reaching the target and penalize time steps to encourage efficiency. 
This reward design, combined with the collision predictor, allows the second stage training to improve collision-free navigation without over-conservative behavior, balancing safety and task success.

Following previous work \cite{akgvp, hoz, tdanet, ORG, mjol, lstde}, A3C \cite{a3c} reinforcement learning method is adopted to train the navigation agent. Let $\theta$, $\theta_v$, and $\phi$ denote the global shared parameters of the actor, the critic , and the collision prediction module, respectively. Their thread-specific copies are $\theta'$,  $\theta_v'$ and $\phi'$, and $E$ is the global episode counter. The actor network learns a policy $\pi(a_t\mid O_t;\theta)$, which predicts the action $a_t$ based on the observation $O_t$. The critic network estimates the state-value function $V(O_t;\theta_v)$ based on the observation $O_t$. The collision prediction module predicts collision probability $P(I_t;\phi)$ using the current RGB image $I_t$. Algorithm~\ref{algorithm} summarizes the per-thread workflow of the proposed two-stage training method.
During the first stage ($E < E_1$), the agent explores freely while the collision prediction module is updated using \texttt{MoveAhead} actions. During the second stage ($E_1 \le E < E_1 + E_2$), the learned predictor guides navigation under a collision penalty, while its parameters remain fixed to prevent overfitting. This design ensures effective exploration in the first stage and safe, collision-free navigation in the second.
\section{Experiments}

We first evaluate the proposed framework across multiple navigation models, followed by comparisons with existing collision avoidance strategies, ablation studies to isolate key components, and finally real-world validation to assess sim-to-real generalization.

\subsection{Experimental Setup}
We evaluate the proposed two-stage training method on selected navigation models and compare it to conventional collision avoidance strategies in the AI2-THOR environment\cite{ai2thor}, which consists of 120 near photo-realistic indoor room scenes spanning four room types, i.e., kitchen, living room, bedroom and bathroom. Each navigation model is trained for 3 million episodes using 32 asynchronous agents with 1 million episodes in the first training stage ($E_1=1,000,000$) and 2 million episodes in the second training stage ($E_2=2,000,000$). The model parameters are optimized with Adam using a learning rate of 0.0001.
All experiments are repeated over five independent trials, and the results are presented as mean$\pm$standard deviation. We use an upward arrow $\uparrow$ to indicate the increase in the CF-SR/CF-SPL after deploying the proposed training method.

\subsection{Benchmark Evaluation Across Navigation Models}

We deploy the proposed training method on the following navigation models. {\bf Baseline}\cite{baseline} concatenates the image features with the word embedding or class label of the target object as the input of the navigation policy network. {\bf HOZ}\cite{hoz} utilizes hierarchical object-to-zone graph to guide the agent. {\bf L-sTDE}\cite{lstde} calculates the object layout gap between different scenes to adjust the prediction of the navigation policy. {\bf AKGVP}\cite{akgvp} leverages the language-image pertaining model to align knowledge graph with visual perception. {\bf MJOLNIR-r}\cite{mjol} utilizes context vectors and the graph convolutional neural network to learn parent-target hierarchical object relationships. {\bf TDANet}\cite{tdanet} learns spatial and semantic relationships of objects through a target attention module and the Siamese network design; image features encoded by the CLIP\cite{CLIP} model are added for collision prediction.

\begin{table}[tbp]
\caption{CF-SR and CF-SPL of different models in AI2-THOR.}
\renewcommand{\arraystretch}{1.3}
\begin{center}
\begin{tabular}{@{}c|ll@{}l>{\hspace*{1.8em}}l@{}>{\hspace*{-0.5em}}l}
\toprule
Data & Model & CF-SR(\%) && CF-SPL(\%) \\
\hline
\multirow{8}{*}{\rotatebox[origin=c]{90}{Offline Data 1\cite{akgvp}}} 
&Baseline\cite{baseline} & 55.8{\tiny $\pm$2.7} && 32.2{\tiny $\pm$1.8} &\\
&HOZ\cite{hoz} & 58.1{\tiny $\pm$0.9} && 33.2{\tiny $\pm$0.5} &\\
&L-sTDE\cite{lstde} & 60.0{\tiny $\pm$0.4} &&34.0{\tiny $\pm$1.1}& \\
&AKGVP\cite{akgvp} & 70.5{\tiny $\pm$0.7}&& 40.4{\tiny $\pm$0.8} \\
&{\bf ours}(Baseline) &59.3{\tiny $\pm$1.0} & {\bf 3.5$\uparrow$} & 33.1{\tiny $\pm$1.5} & {\bf 0.9$\uparrow$}\\
&{\bf ours}(HOZ) &60.2{\tiny $\pm$1.3} & {\bf 2.1$\uparrow$} & 34.1{\tiny $\pm$0.7} & {\bf 0.9$\uparrow$}\\
&{\bf ours}(L-sTDE) & 65.3{\tiny $\pm$0.6} & {\bf 5.3$\uparrow$}& 35.9{\tiny $\pm$0.9} & {\bf 1.9$\uparrow$}\\
&{\bf ours}(AKGVP) &74.4{\tiny $\pm$0.7} & {\bf 3.9$\uparrow$} & 41.6{\tiny $\pm$0.5} & {\bf 1.2$\uparrow$}\\
\hline
\multirow{6}{*}{\rotatebox{90}{Offline Data 2\cite{mjol}}} 
&Baseline\cite{baseline} & 27.1{\tiny $\pm$1.1} && 8.4{\tiny $\pm$0.4}\\
&MJOLNIR-r\cite{mjol} & 47.2{\tiny $\pm$4.1} && 17.9{\tiny $\pm$2.1}\\
&TDANet\cite{tdanet} & 53.5{\tiny $\pm$1.5} && 23.4{\tiny $\pm$1.2}\\
&{\bf ours}(Baseline) & 30.1{\tiny $\pm$0.7} & {\bf 3.0$\uparrow$} & 9.5{\tiny $\pm$0.1} & {\bf 1.1$\uparrow$} \\
&{\bf ours}(MJOLNIR-r) & 55.8{\tiny $\pm$5.0} & {\bf 8.5$\uparrow$} & 21.0{\tiny $\pm$2.3} & {\bf 3.1$\uparrow$}\\
&{\bf ours}(TDANet) & 60.5{\tiny $\pm$2.3} & {\bf 7.0$\uparrow$} & 25.8{\tiny $\pm$0.7} & {\bf 2.4$\uparrow$}\\
\bottomrule
\end{tabular}
\label{tab_comp}
\end{center}
\end{table}

Two offline datasets in the AI2THOR environment are used, which vary in target objects and labels of object detection. The setting of Offline Data 1 follows \cite{akgvp} with 80 rooms as the training set and 20 rooms as the test set. The setting of Offline Data 2 follows \cite{mjol} with 80 rooms as the training set and 40 rooms as the test set. 

The evaluation results of different models using the proposed training method are presented in Table~\ref{tab_comp}. It is observed that the proposed method improves navigation performance consistently with higher CF-SR and CF-SPL for all the selected models. Specifically, in Offline Data 1, L-sTDE and AKGVP using the proposed method increase CF-SR by 5.3\% and 3.9\%, and CF-SPL by 1.9\% and 1.2\%, respectively. In Offline Data 2, MJOLNIR-r and TDANet increase CF-SR by 8.5\% and 7.0\%, and CF-SPL by 3.1\% and 2.4\%, respectively.

\begin{figure}[t]
\centerline{\includegraphics[width=1.0\linewidth]{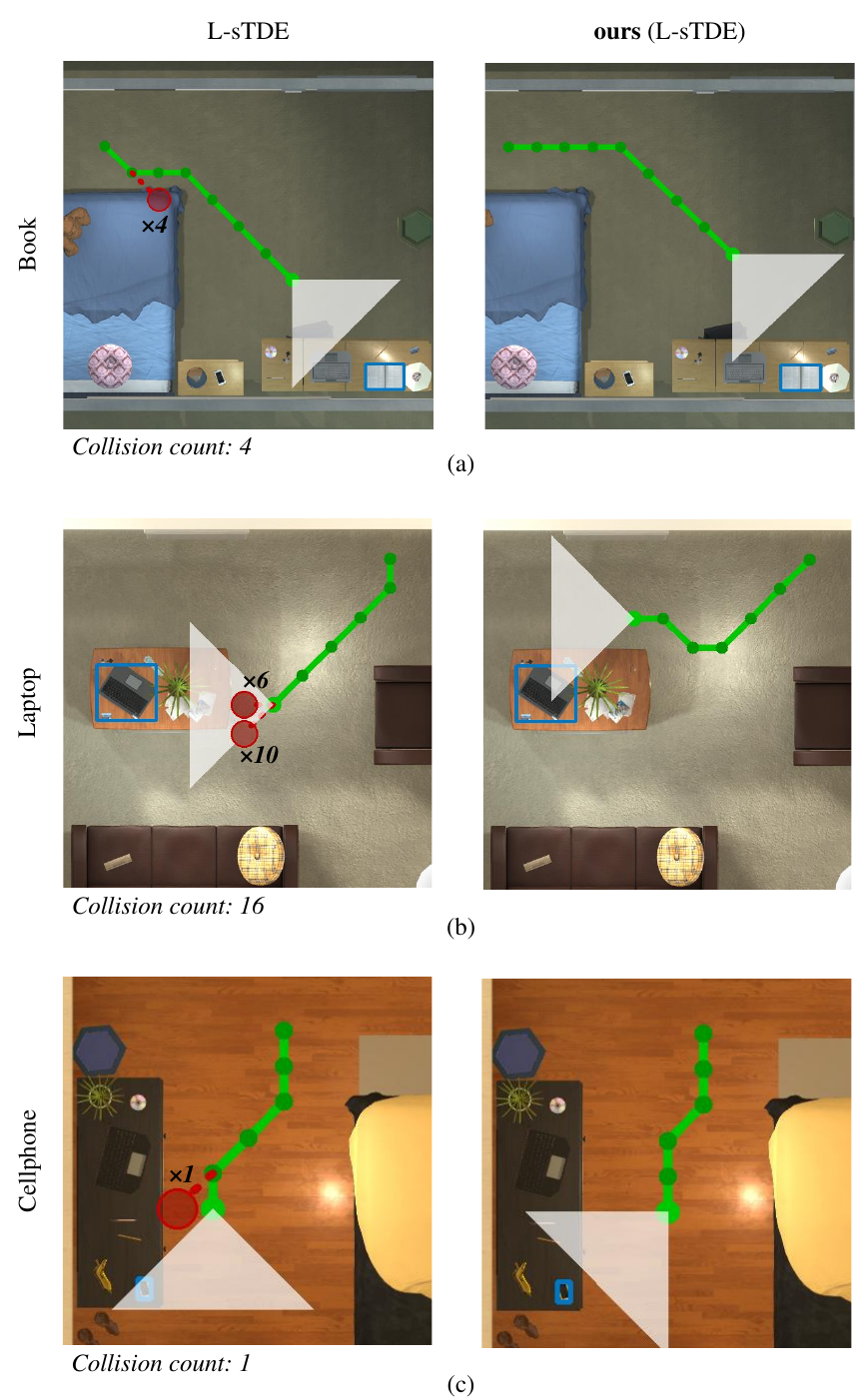}}
\caption{Representative navigation paths of L-sTDE\cite{lstde} before and after applying the proposed training method. Red circles indicate collision-causing actions, with $\times N$ denoting repeated failures at the same location. Blue bounding boxes denote target objects.}
\label{img_path}
\end{figure}

Figure~\ref{img_path} illustrates representative navigation paths of L-sTDE model before and after applying the proposed training method. With the proposed method, L-sTDE successfully bypasses obstacles while maintaining safe distances. For example, in Fig.~\ref{img_path}(b), the original L-sTDE agent becomes stuck in front of a desk, whereas the agent trained with the proposed method rounds the corner of the table and progresses towards the target efficiently.

\subsection{Comparison With Collision Avoidance Methods}\label{sec_comp}

To evaluate the advantages of the proposed method, we adopt L-sTDE trained on Offline Data 1 as the baseline and compare it with several collision avoidance approaches. Specifically, the following methods are considered. {\bf Reward} adds a negative reward $r_{\rm c}=-0.1$ when a collision occurs. {\bf Xiao et al.}\cite{coll1} combines the collision reward $r_{\rm c}=-0.1$ with a single-step shaping reward $r_s=\lambda(d_{t-1}-d_{t})$, where $d_{t-1}$ and $d_t$ denote the path length from the agent to the target at consecutive time steps $t-1$ and $t$ and $\lambda$ is a scaling factor set to 0.01. {\bf Wu et al.}\cite{dual} jointly optimizes the navigation policy with a collision prediction network by sharing the same image feature encoder. 

\begin{figure}[t]
\centerline{\includegraphics[width=1.0\linewidth]{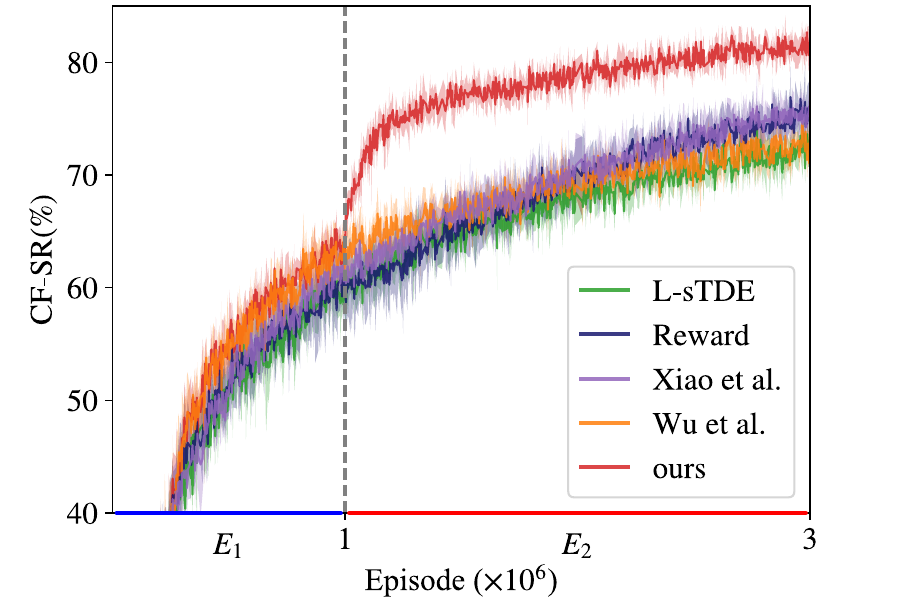}}
\caption{Comparison of CF-SR learning curves under different collision avoidance methods.}
\label{learning_curve}
\end{figure}

The evaluation results on the AI2-THOR test set  are presented in Table~\ref{tab_comp2}. The proposed method achieves the highest  CF-SR of 65.3\% (5.3\%$\uparrow$) and CF-SPL of 35.9\% (1.9\%$\uparrow$). In contrast, the alternative methods exhibit only marginal improvement in CF-SR ($< 2\%$$\uparrow$) and CF-SPL ($< 1\%$$\uparrow$). Figure~\ref{learning_curve} illustrates the CF-SR learning curves during training. 

During the first training stage ($E_1$), the agent simultaneously learns target-seeking and collision prediction without incurring a collision penalty,  resulting in CF-SR increases comparable to Wu et al. In the second stage, the trained collision predictor guides navigation under a collision penalty, enabling rapid improvement in CF-SR to approximately 80\%, as illustrated in Fig.~\ref{learning_curve}. By contrast, the Reward and Xiao et al. methods, which rely solely on reward shaping, induce conservative navigation policies and yield limited CF-SR improvements in both training and test sets. These results demonstrate that the proposed two-stage training framework with collision prediction achieves the most substantial enhancement in collision-free navigation performance among the methods evaluated.

\begin{table}[tbp]
\caption{CF-SR and CF-SPL of different collision avoidance methods in AI2-THOR.}
\renewcommand{\arraystretch}{1.3}
\begin{center}
\begin{tabular}{ll@{}l>{\hspace*{1.8em}}l@{}>{\hspace*{-0.5em}}l}
\toprule
Method & CF-SR(\%) && CF-SPL(\%) \\
\hline
L-sTDE\cite{lstde} & 60.0{\tiny $\pm$0.4} &&34.0{\tiny $\pm$1.1}& \\
Reward &61.2{\tiny $\pm$2.3} & 1.2$\uparrow$ & 34.2{\tiny $\pm$1.2} &  0.2$\uparrow$\\
Xiao et al.\cite{coll1} & 61.5{\tiny $\pm$1.7}& 1.5$\uparrow$& 34.2{\tiny $\pm$1.4} &  0.2$\uparrow$ \\
Wu et al.\cite{dual} &61.6{\tiny $\pm$1.7} & 1.6$\uparrow$ & 34.4{\tiny $\pm$0.5} &  0.4$\uparrow$\\
{\bf ours} & 65.3{\tiny $\pm$0.6} & {\bf 5.3$\uparrow$}& 35.9{\tiny $\pm$0.9} & {\bf 1.9$\uparrow$}\\
\bottomrule
\end{tabular}
\label{tab_comp2}
\end{center}
\end{table}

\subsection{Ablation Study}

Following the experimental setting in Section~\ref{sec_comp}, ablation studies are conducted to isolate the contributions of the two key components: the two-stage training strategy and the collision prediction module. Specifically, we consider two comparison counterparts: Stage Only, which removes the collision prediction module and retains collision reward $r_c=-0.5$ in the second training stage, and CP Only, which updates the collision prediction module and feed the predicted collision probability into the navigation policy throughout the training process without incorporating the collision reward. 

\begin{figure}[t]
\centerline{\includegraphics[width=1.0\linewidth]{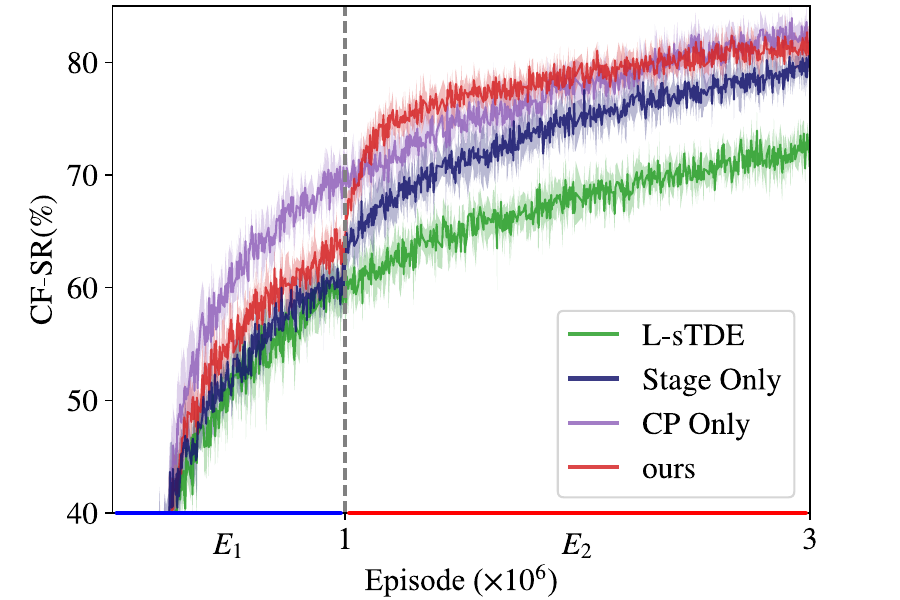}}
\caption{CF-SR learning curves for ablation studies on the training set.}
\label{learning_ablation}
\end{figure}

The CF-SR and CF-SPL for the ablation studies are presented in Table~\ref{tab_ablation} and Fig.~\ref{learning_ablation}. Results indicate that the two-stage training contributes the most to the overall performance, increasing the CF-SR and CF-SPL in the test set by 3.3\% and 1.3\%, respectively. With the two-stage training method, the agent first acquires effective target-seeking behavior and subsequently learns collision avoidance, thereby avoiding the overly conservative navigation policies that emerge when a collision penalty is directly applied.

While the CP Only method reaches the highest CF-SR in the training set, its performance gains in the test set are minimal. We conjecture that this discrepancy arises from overfitting. Continuous updating of the collision prediction module during training reduces exploration and alters navigation trajectories, as collisions become less frequent in later episodes. As illustrated in Fig.~\ref{learning_ablation}, the CF-SR of  CP Only  exceeds 80\% during training, however, its test performance deteriorates because the navigation policy relies heavily on the collision prediction, which are nearly perfect in  training  but less accurate in unseen environments. In contrast, the proposed method restricts collision prediction training to the first stage and incorporates it in the navigation policy alongside the collision reward during the second stage. This design mitigates overfitting and enhances generalization. The ablation study results confirm the effectiveness of the two-stage collision-aware training framework.

\begin{table}[t]
\caption{Ablation study results on the AI2-THOR test set.}
\renewcommand{\arraystretch}{1.3}
\begin{center}
\begin{tabular}{ll@{}l>{\hspace*{1.8em}}l@{}>{\hspace*{-0.5em}}l}
\toprule
Method & CF-SR(\%) && CF-SPL(\%) \\
\hline
L-sTDE\cite{lstde} & 60.0{\tiny $\pm$0.4} &&34.0{\tiny $\pm$1.1}& \\
Stage Only &63.3{\tiny $\pm$0.5} & 3.3$\uparrow$ & 35.3{\tiny $\pm$0.2} &  1.3$\uparrow$\\
CP Only & 60.2{\tiny $\pm$1.3} &  0.3$\uparrow$& 33.5{\tiny $\pm$1.3} &  0.5$\downarrow$\\
{\bf ours} & 65.3{\tiny $\pm$0.6} & {\bf 5.3$\uparrow$}& 35.9{\tiny $\pm$0.9} & {\bf 1.9$\uparrow$}\\
\bottomrule
\end{tabular}
\label{tab_ablation}
\end{center}
\end{table}

\subsection{Real-World Evaluation and Zero-Shot Deployment}

To validate the proposed collision-aware training framework beyond simulation, real-world experiments are conducted on a TurtleBot4 mobile robot in cluttered indoor environments. Compared to simulation, real-world deployment introduces additional challenges, including sensor noise, visual domain shift, and complex obstacle distributions. These factors make reliable collision-aware navigation critical for safe and effective operation.

TDANet \cite{tdanet} is adopted as the underlying navigation policy  due to its demonstrated zero-shot generalization capability. The model is trained entirely in simulation and directly deployed in real-world environments without fine-tuning, providing a strict evaluation of sim-to-real transfer.

To mitigate perception instability under domain shift, LiDAR-based collision sensing (RPLIDAR A1M8) is employed during deployment. This design isolates the effect of collision-aware decision-making from RGB perception errors while preserving the original policy structure.

The robotic platform is equipped with an OAK-D Pro RGB camera mounted at 1.3 m with active viewpoint control. All models are executed in real time on an onboard computing system with an i7-14700KF CPU and an NVIDIA GeForce RTX 4060 Ti GPU. Evaluation is conducted across three indoor environments with varying layouts and obstacle densities, with five trials per scene with a total of 15 trials.

As summarized in Table~\ref{tab_exp}, the baseline TDANet achieves 0/15 success (0.0\% CF-SR), indicating frequent failure in real-world environment due to insufficient collision awareness. In contrast, the proposed method achieves 12/15 successes (80.0\% CF-SR), demonstrating a substantial improvement in deployment robustness. Notably, this corresponds to an absolute gain of 80.0\% over the baseline, highlighting the effectiveness of incorporating collision-aware reasoning for sim-to-real transfer.

\subsection{Real-World Ablation Study}

To quantify the contribution of each component under real-world conditions, an ablation study is conducted by comparing the full  framework with two variants: (1) {Reward Only}, which employs a single-stage training strategy with collision penalties,  and (2) {Stage Only}, which retains the two-stage training strategy but removes collision prediction from the policy input. This design isolates the effects of staged training and explicit collision modeling.

\begin{figure*}[t]
\centerline{\includegraphics[width=1.0\linewidth]{trial2.0.pdf}}
\caption{Real-world experiment visualization. Target objects include a potted plant, a book, and a laptop. Each row corresponds to one of the four compared methods. Trajectories are encoded with a light-to-dark color gradient over time. The proposed method generates more efficient and collision-free paths, whereas others suffer from oscillations, longer paths, or collisions.}
\label{test}
\end{figure*}

\begin{table}[t]
\caption{Real-world evaluation and ablation results over 15 trials.}
\renewcommand{\arraystretch}{1.3}
\begin{center}
\begin{tabular}{lcc}
\toprule
Method & Succ Count & CF-SR (\%) \\
\hline
Baseline     & 0/15  & 0.0 \\
Reward Only  & 1/15  & 6.7 \\
Stage Only   & 4/15  & 26.7 \\
{\bf Ours}   & 12/15 & 80.0 \\
\bottomrule
\end{tabular}
\label{tab_exp}
\end{center}
\end{table}

As shown in Table~\ref{tab_exp}, performance improves consistently with each component. Reward Only achieves 1/15 success (6.7\% CF-SR), indicating that collision penalties alone are insufficient to prevent failure in complex real-world layouts and may lead to overly conservative behavior. Stage Only improves performance to 4/15 (26.7\% CF-SR), suggesting that decoupling exploration and safety learning stabilizes training but remains fundamentally reactive without explicit collision anticipation.

The full framework achieves the best performance with 12/15 successes (80.0\% CF-SR), significantly outperforming both variants. This improvement demonstrates that integrating staged training with collision prediction enables proactive avoidance behavior rather than reactive correction. Consequently, the agent maintains higher success rates in dense obstacle configurations where delayed reactions typically lead to failure.

Representative results are illustrated in Fig.~\ref{test}. Reward Only exhibits oscillations near obstacles, while Stage Only improves smoothness but still incurs collisions. The proposed method produces efficient and collision-free trajectories. Overall, the results confirm that both staged training and collision prediction are essential for robust real-world navigation.


\section{Conclusion}

This letter introduced a collision-aware evaluation metric CF-SR and adopted CF-SPL to evaluate object-goal visual navigation performance under collision constraints. A two-stage DRL training framework with explicit collision prediction was proposed to improve the collision-free navigation performance while maintaining effective target-reaching ability. 
The proposed framework was deployed across multiple navigation models and evaluated against existing collision avoidance methods in the AI2-THOR environment. Experimental results demonstrated consistent improvements in both CF-SR and CF-SPL, highlighting the effectiveness and general applicability of the proposed method. Ablation studies further validated the effectiveness of the two-stage training strategy and the collision prediction module. In addition, real-world experiments confirmed the superior performance and practical applicability of the proposed method in cluttered environments.

Future work will focus on improving RGB-based collision prediction to further reduce the sim-to-real gap and enhance robustness in real-world deployments.It would also be valuable to explore more effective integration of semantic understanding and collision prediction to further improve navigation performance.

\bibliographystyle{IEEEtran}
\bibliography{IEEEabrv,mybibfile}
\end{document}